\documentclass[10pt,twocolumn,letterpaper]{article}

\usepackage{iccv}              % To produce the CAMERA-READY version
\usepackage[accsupp]{axessibility}

\usepackage{pifont}

\definecolor{iccvblue}{rgb}{0.21,0.49,0.74}
\usepackage[pagebackref,breaklinks,colorlinks,allcolors=iccvblue]{hyperref}
\usepackage{orcidlink}

\def\paperID{*****} % *** Enter the Paper ID here
\def\confName{ICCV}
\def\confYear{2025}

\title{TinyGiantVLM: A Lightweight Vision-Language Architecture for Spatial Reasoning under Resource Constraints}

\renewcommand{\and}{\hspace{2em}} % Chỉnh để tên của các tác giả ở cùng 1 hàng

\author{
Vinh-Thuan Ly~\orcidlink{0009-0005-7557-8231} \and
Hoang M. Truong~\orcidlink{0009-0008-3899-895X} \and
Xuan-Huong Nguyen~\orcidlink{0009-0008-3902-5375} \\
University of Science, VNU-HCM, Ho Chi Minh City, Vietnam \\
Vietnam National University, Ho Chi Minh City, Vietnam \\
{\tt\small \{22280092, 22280034, 22280037\}@student.hcmus.edu.vn}\\
\href{https://tinygiantvlm.github.io/}{\small{\texttt{https://tinygiantvlm.github.io/}}}
}
\begin{document}
\maketitle
\begin{abstract}

Reasoning about fine-grained spatial relationships in warehouse-scale environments poses a significant challenge for existing vision-language models (VLMs), which often struggle to comprehend 3D layouts, object arrangements, and multimodal cues in real-world industrial settings. In this paper, we present TinyGiantVLM, a lightweight and modular two-stage framework designed for physical spatial reasoning, distinguishing itself from traditional geographic reasoning in complex logistics scenes. Our approach encodes both global and region-level features from RGB and depth modalities using pretrained visual backbones. To effectively handle the complexity of high-modality inputs and diverse question types, we incorporate a Mixture-of-Experts (MoE) fusion module, which dynamically combines spatial representations to support downstream reasoning tasks and improve convergence. Training is conducted in a two-phase strategy: the first phase focuses on generating free-form answers to enhance spatial reasoning ability, while the second phase uses normalized answers for evaluation. Evaluated on Track 3 of the AI City Challenge 2025, our 64M-parameter base model achieved 5th place on the leaderboard with a score of 66.8861, demonstrating strong performance in bridging visual perception and spatial understanding in industrial environments. We further present an 80M-parameter variant with expanded MoE capacity, which demonstrates improved performance on spatial reasoning tasks.

\end{abstract}    
\section{Introduction}
\label{sec:intro}

Understanding fine-grained spatial relationships in industrial environments presents a significant challenge for artificial intelligence. While AI systems have achieved impressive performance in image-based recognition and short-form spatial tasks, they often struggle to reason about 3D object layouts, dimensions, and spatial relations in real-world, logistics-scale environments. Prior research in visual question answering and 3D scene understanding has largely focused on synthetic datasets or domestic scenes~\cite{Khanna_2024_CVPR_SyntheticScene, zhang-etal-2024-countercurate, robopoint2024}, leaving a substantial gap in industrial settings such as warehouses and logistics hubs.

\begin{figure}[t]
    \centering
    \includegraphics[width=1\linewidth]{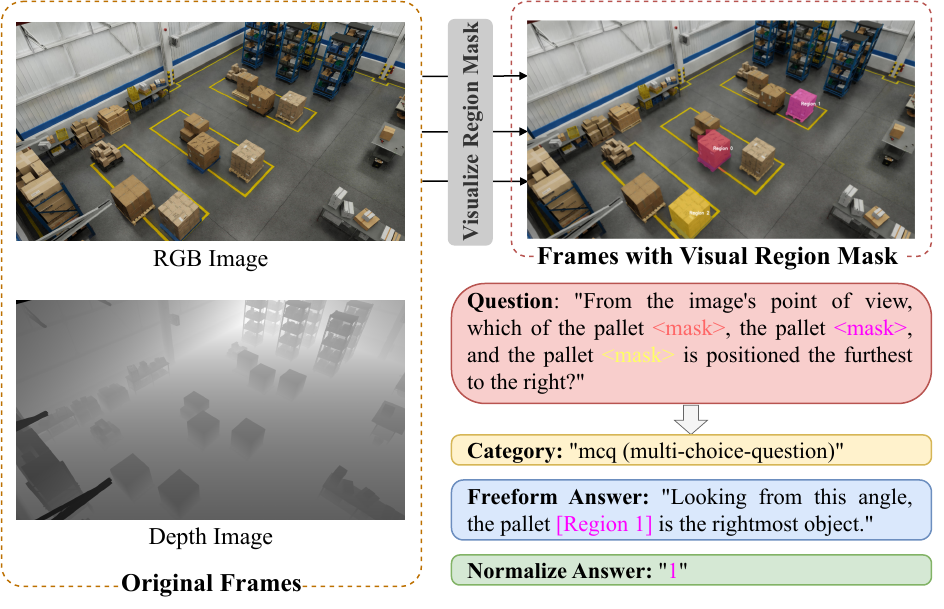}
    \vspace{-20pt}
    \caption{An example of multiple-choice question (MCQ) in the dataset}
    \label{fig:intro}
    \vspace{-10pt}
\end{figure}

The AI City Challenge \cite{Tang25AICity25} Track 3: Warehouse Spatial Intelligence addresses this gap using the PhysicalAI-Spatial-Intelligence-Warehouse dataset for warehouse-scale 3D scene understanding through natural language questions. The challenge encompasses four distinct types of spatial reasoning tasks: distance estimation, object counting, multiple-choice questions (MCQ) for spatial grounding, and left/right spatial relation queries. These tasks demand AI systems capable of integrating visual perception, geometric reasoning, and language comprehension while operating under resource constraints typical of industrial deployment scenarios. \Cref{fig:intro} illustrates an example of a multiple-choice question (MCQ) from the dataset.

To address these challenges, we propose \textbf{TinyGiantVLM}, a lightweight multimodal encoder-decoder architecture designed for precise spatial reasoning under resource constraints. Our approach processes both RGB and depth information at global and region levels through a dual-branch design, integrating cross-attention fusion mechanisms with a Mixture-of-Experts (MoE) framework based on FuseMoE~\cite{fusemoe}. The architecture employs specialized expert networks for each of the four question types, enabling task-specific reasoning while maintaining computational efficiency through sparse expert routing.

Unlike traditional geographic reasoning models that rely on symbolic or coordinate-based knowledge (e.g., "Is Beijing north of Shanghai"), our focus is on physical spatial reasoning through visual perception, learning relative object positioning and geometric relationships (e.g., "Which pallet is furthest right") from RGB-D inputs. This distinction emphasizes the need for models that ground spatial language in perceptual understanding rather than abstract knowledge bases.

Our key contributions are as follows:
\begin{itemize}
    \item We introduce TinyGiantVLM, a resource-efficient vision-language architecture that achieves strong spatial reasoning performance while maintaining low computational overhead suitable for industrial deployment.
    \item We propose a dual-branch multimodal feature extraction approach that effectively combines RGB and depth information at both global and region levels.
    % \item We demonstrate the effectiveness of sparse MoE fusion for task-specific spatial reasoning across the four question types (distance, count, MCQ, and left/right).
    \item Through extensive experiments, we validate our approach on the AI City Challenge \cite{Tang25AICity25} Track 3, achieving competitive results with significantly reduced computational requirements.
\end{itemize}

In \cref{sec:related_work}, we provide a brief review of existing methods related to our problem and solution approach. We then present the details of our proposed framework in \cref{sec:method}. The experimental setup and evaluation results are discussed in \cref{sec:experiment}, followed by concluding remarks, future directions and limitations in \cref{sec:conclusion}.

\section{Related Work}
\label{sec:related_work}

\paragraph{Spatial Vision Language Models.}
Recent works have focused on addressing spatial problems through the reasoning capabilities of LLMs, leading to the development of various 3D vision-language models with distinct input modalities and alignment strategies. Image-based methods derive 3D understanding from 2D images~\cite{zhu2024llava, zhang2024agent3d, qi2024shapellm, SpatialVLM_Deepmind}, with approaches like SpatialVLM~\cite{SpatialVLM_Deepmind} enhancing spatial reasoning through Internet-scale 3D spatial reasoning data with quantitative metric capabilities. Point cloud-based approaches employ different alignment strategies~\cite{xu2025pointllm, huang2024chat, qi2024gpt4point, tang2024minigpt, zhu2024scanreason, wang2023chat, zhu2023pointclip, zhang2022pointclip}, ranging from direct alignment methods like PointLLM~\cite{xu2025pointllm} to task-specific methods like ScanReason~\cite{zhu2024scanreason} for reasoning grounding. Hybrid approaches combine multiple modalities~\cite{guo2023point, zhou2023uni3d, hong2024multiply}, such as Point-bind~\cite{guo2023point} which integrates point clouds and images for cross-modal understanding. More recently, SpatialRGPT~\cite{cheng2024spatialrgpt} introduces regional representation learning from 3D scene graphs with flexible depth integration modules, demonstrating strong generalization in spatial reasoning tasks and robotic applications.

However, most of these methods heavily rely on large-scale language models (LLMs), making them computationally expensive and less suitable for low-resource environments. In contrast, our work focuses on enabling spatial vision-language reasoning with reduced computational requirements.

\paragraph{Language and Visual Backbones.}
Pretrained language and vision backbones have proven essential for multimodal reasoning tasks. The Text-to-Text Transfer Transformer (T5)~\cite{JMLR:v21:20-074} is a unified encoder-decoder model that reformulates diverse NLP tasks into a text-to-text format, enabling broad generalization. For vision, CLIP-ViT~\cite{pmlr-v139-radford21a} leverages contrastive learning on image-text pairs to produce global semantic embeddings, while DPT~\cite{ranftl2021visiontransformersdenseprediction} introduces dense spatial features for monocular depth estimation. These models have been widely adopted as backbones in multimodal systems, where global and region-level visual features are combined with pretrained language representations to support complex spatial understanding.

%\paragraph{Region-level Visual Language Models.}

\paragraph{Mixture-of-Experts.}
The Mixture-of-Experts (MoE) framework was first introduced by Jordan \etal~\cite{jordan1994moe}, establishing a theoretical foundation for modular learning with conditional computation. Shazeer \etal~\cite{google2017moe} scaled MoE to deep neural networks using softmax-based sparse gating, enabling parameter efficiency and specialization. More recently, MoE has been adapted to vision-language tasks and multimodal reasoning, where semantic alignment is critical. SparseMoE~\cite{li2023sparsemixtureofexpertsdomaingeneralizable} proposed cosine similarity as a gating function to encourage expert selection based on domain similarity. FuseMoE~\cite{fusemoe} introduced a novel Laplace-based gating mechanism, theoretically shown to improve convergence and empirically demonstrated to enhance expert utilization and predictive performance in heterogeneous, multimodal settings.
\section{Methodology}
\label{sec:method}

\begin{figure*}
    \centering
    \includegraphics[width=1\linewidth]{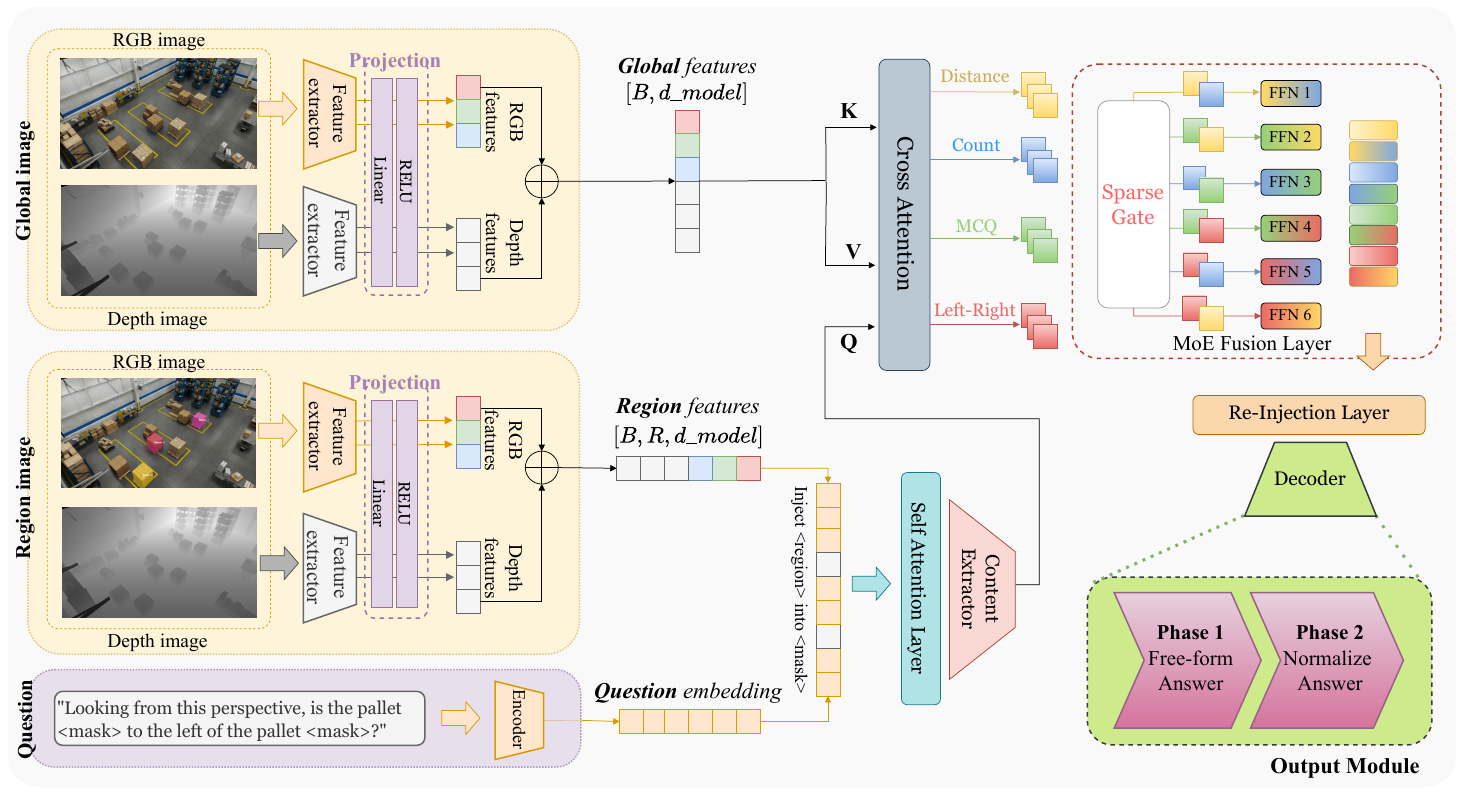}
    \vspace{-20pt}
    \caption{Proposed architecture of TinyGiantVLM. The model processes both global and region-level RGB and depth images through feature extractors, followed by projection and fusion of RGB-depth features. The resulting global and region features are integrated using cross-attention with question embeddings. Region features are injected into masked positions in the question and passed through a self-attention layer and content extractor. Task-specific reasoning is designed to be handled by a Mixture-of-Experts (MoE) fusion layer guided by a sparse gate, with separate experts for different question types. While this MoE module is part of our proposed design, it was not activated in the final evaluation due to implementation issues. The fused representation is then re-injected into the decoder to generate free-form answers. In Phase 2, the model is fine-tuned on normalized answers using the same input questions as in Phase 1, allowing it to retain spatial reasoning ability while adapting to structured output formats.}
    \vspace{-5pt}
    \label{fig:architecture}
\end{figure*}

This section outlines the methodological framework of TinyGiantVLM, detailing how global and region-level visual features are extracted from RGB and depth modalities (\cref{subsec:visual_feature_extraction}) and integrated with language components through cross-attention fusion and task-specific expert routing (\cref{subsec:TinyGiantVLM}) to enable efficient spatial reasoning in industrial scenarios. \Cref{fig:architecture} illustrates our overall architecture.

\subsection{Visual Feature Extraction}
\label{subsec:visual_feature_extraction}

\paragraph{Global Visual Feature Extraction.}
To capture both semantic and geometric cues from the input image, we employ two complementary pretrained vision transformers. Specifically, we use the OpenAI CLIP-ViT-Large-Patch14 model~\cite{pmlr-v139-radford21a} for the RGB modality and the Intel DPT-Hybrid-MiDaS model~\cite{ranftl2021visiontransformersdenseprediction} for the depth modality. For each modality, we extract a global visual feature by taking the output embedding of the \texttt{[CLS]} token from the final encoder layer. These compact feature representations from both RGB and depth branches are subsequently fused and utilized in downstream multi-modal reasoning tasks.
\begin{equation}
    \mathbf{f}^{\mathrm{RGB}} = \mathbf{e}^{\mathrm{RGB}}_{[\mathrm{CLS}]}, \quad 
    \mathbf{f}^{\mathrm{Depth}} = \mathbf{e}^{\mathrm{Depth}}_{[\mathrm{CLS}]}
    \label{eq:global_rgb_depth}
\end{equation}

\paragraph{Region-level Feature Extraction.}

To obtain region-level representations from both RGB and depth modalities, we leverage the patch embeddings produced by the last layer of each vision transformer branch. Let $I \in \mathbb{R}^{3 \times H \times W}$ denote the input RGB image and $D \in \mathbb{R}^{3 \times H' \times W'}$ the input depth map (with the original single-channel depth map repeated across three channels). For CLIP-ViT, images are resized to $224 \times 224$ and split into non-overlapping patches of size $14 \times 14$, resulting in a $16 \times 16$ grid. For DPT, depth images are resized to $384 \times 384$ with a patch size of $16 \times 16$, yielding a $24 \times 24$ grid.

Each region is annotated using a segmentation mask, provided in run-length encoding (RLE) format. We first decode each RLE mask into a binary mask matching the original image resolution, and then downsample the mask to the corresponding patch grid resolution for each modality ($16 \times 16$ for RGB, $24 \times 24$ for depth).

Given $N$ patch embeddings $\mathbf{E} = [\mathbf{e}_1, \ldots, \mathbf{e}_N]$ from CLIP-ViT (where $N = 16 \times 16$), and $N'$ patch embeddings $\mathbf{E}' = [\mathbf{e}'_1, \ldots, \mathbf{e}'_{N'}]$ from DPT (where $N' = 24 \times 24$), we define a binary region mask $M_r \in \{0,1\}^{G \times G}$ or $M'_r \in \{0,1\}^{G' \times G'}$ for each region $r$ in the image, downsampled to match the spatial layout of patches in each branch. The region feature for region $r$ in the RGB branch is computed as:
\begin{equation}
    \mathbf{f}_r^{\mathrm{RGB}} = \frac{1}{|P_r|} \sum_{i \in P_r} \mathbf{e}_i
\end{equation}
where $P_r$ is the set of patch indices whose spatial locations fall within the mask $M_r$ (i.e., $M_r(i) = 1$). Similarly, for the depth branch:
\begin{equation}
    \mathbf{f}_r^{\mathrm{Depth}} = \frac{1}{|P'_r|} \sum_{i \in P'_r} \mathbf{e}'_i
\end{equation}
where $P'_r$ is defined by the corresponding mask $M'_r$ over the depth patch grid.

To reduce computation and enable faster training, we pre-extract and cache both global and region-level features using pretrained transformer branches. These cached features are stored on disk and reused during downstream model training.

\subsection{TinyGiantVLM}
\label{subsec:TinyGiantVLM}
Our TinyGiantVLM is a region-aware, dual-branch multimodal encoder-decoder architecture that integrates both global and region-level visual features with flexible language grounding, as illustrated in Figure~\ref{fig:architecture}. The core model is based on the T5-small encoder-decoder backbone~\cite{JMLR:v21:20-074}, which we substantially extend with (i) dual RGB-depth visual branches, (ii) multimodal region feature injection, (iii) a cross-attention fusion module for fine-grained visual-language alignment, and (iv) a Mixture-of-Experts (MoE) fusion layer designed for adaptive, task-specific reasoning in multimodal scenarios. This modular design enables TinyGiantVLM to support robust and extensible spatial reasoning beyond standard text generation.

\paragraph{Global Visual Branch.}
We reuse the global RGB and depth features extracted in Section~\ref{subsec:visual_feature_extraction}, denoted as $\mathbf{f}^{\mathrm{RGB}}$ and $\mathbf{f}^{\mathrm{Depth}}$. These features are projected into a shared embedding space via modality-specific linear layers, and then concatenated to form the fused global representation:
\begin{align}
    \tilde{\mathbf{f}}^{\mathrm{RGB}} &= W^{\mathrm{RGB}} \mathbf{f}^{\mathrm{RGB}} + \mathbf{b}^{\mathrm{RGB}} \\
    \tilde{\mathbf{f}}^{\mathrm{Depth}} &= W^{\mathrm{Depth}} \mathbf{f}^{\mathrm{Depth}} + \mathbf{b}^{\mathrm{Depth}} \\
    \mathbf{g} &= \left[\tilde{\mathbf{f}}^{\mathrm{RGB}} \; \| \; \tilde{\mathbf{f}}^{\mathrm{Depth}}\right]
\end{align}

\paragraph{Region Branch.}
For each annotated region $j$, we reuse region-level features, denoted as $\mathbf{f}_j^{\mathrm{RGB}}$ and $\mathbf{f}_j^{\mathrm{Depth}}$ for the RGB and depth branches, respectively. These features are projected into a shared embedding space using modality-specific linear layers, resulting in $\tilde{\mathbf{f}}_j^{\mathrm{RGB}}$ and $\tilde{\mathbf{f}}_j^{\mathrm{Depth}}$. The two projected vectors are then concatenated to obtain:
\begin{equation}
    \mathbf{h}_j = \left[\tilde{\mathbf{f}}_j^{\mathrm{RGB}} \, \| \, \tilde{\mathbf{f}}_j^{\mathrm{Depth}}\right]
\end{equation}
Finally, the concatenated vector is passed through a two-layer feed-forward network with ReLU activations:
\begin{equation}
    \mathbf{r}_j = \sigma\left(W_2\, \sigma(W_1\, \mathbf{h}_j + \mathbf{b}_1) + \mathbf{b}_2 \right)
\end{equation}

\paragraph{Question Encoding and Region Injection.}
To align region-level visual features with the input question, we first replace each \texttt{<mask>} token with a unique placeholder token \texttt{<extra\_id\_j>}, where $j$ corresponds to the $j$-th annotated region. The tokenized sequence is then embedded into vectors $\mathbf{T} = [\mathbf{e}_1, \mathbf{e}_2, \ldots, \mathbf{e}_L]$ using the T5 tokenizer.

At each placeholder position $p_j$, we inject the corresponding multimodal region feature $\mathbf{r}_j$ by replacing the token embedding:
\begin{equation}
\mathbf{e}_{p_j} \leftarrow \mathbf{r}_j \quad \text{for } j = 1, \ldots, R
\end{equation}

This results in a modified embedding sequence $\tilde{\mathbf{T}}$, which is fed into the T5 encoder for joint contextualization of natural language and visual region features.

\paragraph{Encoder Contextualization.}
The modified embedding sequence $\tilde{\mathbf{T}}$ is fed into the T5-small encoder~\cite{JMLR:v21:20-074}, which contextualizes both the natural language tokens and the injected region features. The encoder produces contextualized hidden states for all positions:
\begin{equation}
    \mathbf{H} = \operatorname{T5Encoder}(\tilde{\mathbf{T}},\, \mathrm{mask})
\end{equation}
where $\mathbf{H} = [\mathbf{h}_1, \ldots, \mathbf{h}_L]$ denotes the output embeddings for the entire sequence, and $\mathbf{h}_{p_j}$ corresponds to the contextualized representation of the $j$-th region.

\paragraph{Cross-Attention Fusion.}
We extract the contextualized embeddings at the region placeholder positions, denoted as $\mathbf{r}_j^{\mathrm{ctx}} = \mathbf{h}_{p_j}$ for $j = 1, \ldots, R$, where $\mathbf{h}_{p_j}$ is the encoder output at position $p_j$.

Let $\mathbf{R}^{\mathrm{ctx}} = [\mathbf{r}_1^{\mathrm{ctx}}, \ldots, \mathbf{r}_R^{\mathrm{ctx}}]$ be the matrix of contextualized region embeddings, and let $\mathbf{g}$ denote the fused global visual feature. We apply a cross-attention mechanism where each region embedding attends to the global feature:
\begin{align}
    \mathbf{Q} &= \mathbf{R}^{\mathrm{ctx}} W^Q, \quad
    \mathbf{K} = \mathbf{g} W^K, \quad
    \mathbf{V} = \mathbf{g} W^V \\
    \mathbf{C} &= \operatorname{softmax}\left(\frac{\mathbf{Q} \mathbf{K}^\top}{\sqrt{d}}\right)\mathbf{V}
\end{align}
where $\mathbf{C} = [\mathbf{c}_1, \ldots, \mathbf{c}_R]$ are the resulting cross-attended region embeddings. Here, each region query interacts with the global scene context via multi-head attention, enriching its representation with holistic semantics.

\paragraph{MoE Fusion Layer.}
To enable specialization over different spatial reasoning types, we adopt a sparsely-gated Mixture-of-Experts (MoE) fusion layer inspired by FuseMoE~\cite{fusemoe}. Unlike softmax or cosine-based gating~\cite{google2017moe, li2023sparsemixtureofexpertsdomaingeneralizable}, we employ a Laplace-based gating function, which better supports diverse and heterogeneous visual tokens by avoiding norm sensitivity and promoting balanced expert selection.

Given the cross-attended region embeddings $\mathbf{C} = [\mathbf{c}_1,\,\ldots,\,\mathbf{c}_R]$ from the previous step, each embedding $\mathbf{c}_j \in \mathbb{R}^D$ is routed to a subset of experts.

Let $\{E_i\}_{i=1}^S$ be the set of $S$ experts. The router selects the Top-$K$ experts using negative Euclidean distance between $\mathbf{c}_j$ and a set of learned gating vectors $W \in \mathbb{R}^{S \times D}$:
\begin{equation}
    h_\ell(\mathbf{c}_j) = \mathrm{Top~K}(-\|W - \mathbf{c}_j\|_2).
\end{equation}
Let $\mathcal{S}_j$ be the indices of the selected $k$ experts for token $\mathbf{c}_j$. The gating weights are computed using Laplace normalization:
\begin{equation}
    G(\mathbf{c}_j)_i = \frac{\exp(-\|\mathbf{c}_j - W_i\|_2)}{\sum\limits_{l \in \mathcal{S}_j} \exp(-\|\mathbf{c}_j - W_l\|_2)}, \quad i \in \mathcal{S}_j.
    \label{eq:moe_laplace}
\end{equation}
Each expert $E_i: \mathbb{R}^D \rightarrow \mathbb{R}^D$ is a lightweight feed-forward network (FFN). The final output for region $j$ is the weighted sum of the selected expert outputs:
\begin{equation}
    \mathbf{z}_j = \sum_{i \in \mathcal{S}_j} G(\mathbf{c}_j)_i \cdot E_i(\mathbf{c}_j).
\end{equation}

In our setting, we define $S = 4$ experts specialized for four spatial reasoning tasks: (i) distance estimation, (ii) object counting, (iii) multi-choice grounding, and (iv) spatial relation inference. We set $k = 2$ to maintain sparsity while allowing soft routing.

The Laplace gating mechanism in~\cref{eq:moe_laplace} improves convergence and stability. Specifically, unlike inner-product based softmax gating, which can favor experts with high norms and lead to ``representation collapse", our Euclidean distance-based approach promotes a more balanced expert selection by avoiding this inner-product bias~\cite{fusemoe}. This scale-invariant similarity is especially suitable for our multimodal spatial VQA task. In this setting, tokens from different regions and question types are inherently heterogeneous, varying greatly in semantics and magnitude, and our gating mechanism handles these diverse inputs robustly.

\paragraph{Final Re-injection and Answer Generation.}
% The final region representations $\mathbf{z}_j$ from the MoE layer are re-injected into the encoder output at the corresponding region placeholder positions $p_j$, replacing the features from the cross-attention step:
% \begin{equation}
%     \mathbf{h}_{p_j} \leftarrow \mathbf{z}_j \quad \text{for}~j = 1,\ldots,R
% \end{equation}
% This updated sequence of encoder outputs is then passed to the T5-small decoder~\cite{JMLR:v21:20-074}, which generates the final answer or target sequence for the given question. This design supports a variety of downstream tasks such as spatial question answering and scene understanding.

In our design, the final region representations $\mathbf{z}_j$, typically produced by the MoE layer, are re-injected into the encoder output at the corresponding region placeholder positions $p_j$, replacing the features from the cross-attention step:
\begin{equation}
    \mathbf{h}_{p_j} \leftarrow \mathbf{z}_j \quad \text{for}~j = 1,\ldots,R
\end{equation}
% In the current implementation where the MoE module is inactive, we set $\mathbf{z}_j = \mathbf{c}_j$, i.e., the cross-attended embedding from the previous step.

\paragraph{Training Strategy.}

Our training procedure follows a two-phase curriculum designed to balance open-ended spatial reasoning and robust answer normalization. In \textbf{Phase 1}, we train TinyGiantVLM with free-form answers, encouraging the model to develop broad spatial reasoning abilities and generate natural language descriptions that reflect nuanced relationships in the scene. This stage allows the model to learn richer output distributions and fosters a deeper understanding of spatial concepts, unconstrained by rigid answer formats.

Once the model has acquired a strong spatial reasoning prior, we proceed to \textbf{Phase 2}, where it is fine-tuned using normalized answers that match the canonical submission format required by the challenge (e.g., exact numbers, categorical labels such as "left" or "right"). This phase acts as a targeted adaptation, refining the model to prioritize accuracy and consistency in its outputs, while preserving the spatial reasoning skills gained in the initial stage. This curriculum enables TinyGiantVLM to both reason flexibly and generate precise, evaluation-ready predictions.

In both phases, the model is trained to minimize the cross-entropy loss:
\begin{equation}
    \mathcal{L} = - \sum_{t=1}^T \log p(y_t \mid y_{<t},\, \mathbf{x})
\end{equation}
where $y_{1:T}$ is the target answer sequence and $\mathbf{x}$ denotes the input (image, depth, and question).

This two-phase training strategy enables TinyGiantVLM to bridge flexible free-form reasoning with structured prediction, supporting a wide range of industrial visual question answering tasks.

\section{Experiment}
\label{sec:experiment}

\subsection{Dataset}

For fine-tuning and evaluation, We use the PhysicalAI-Spatial-Intelligence-Warehouse dataset from Track 3 of the AI City Challenge 2025 \cite{Tang25AICity25}. Each sample includes paired RGB and monocular depth images, region-level segmentation masks, and spatial reasoning questions covering counting, distance, left/right relation, and multiple-choice grounding. The dataset is generated using Omniverse IsaacSim to simulate diverse industrial layouts with annotated 3D object configurations.

\subsection{Implementation Details}
All experiments are conducted on a single NVIDIA Tesla P100 GPU provided by Kaggle. TinyGiantVLM is fine-tuned in two phases: in Phase 1, the model is trained for 1 epoch with free-form answers; in Phase 2, it is further fine-tuned for 10 epochs with normalized answers. We use the AdamW optimizer with a learning rate of $5 \times 10^{-5}$, weight decay of $1 \times 10^{-2}$, and a batch size of 32. All other hyperparameters follow the default settings of the HuggingFace Transformers library.

To reduce computational cost, we discard the first 120{,}000 samples from the distance category in the training set, preserving the diversity of reasoning types while significantly reducing training time. The non-MoE variant of TinyGiantVLM contains approximately 64M trainable parameters, while enabling the MoE module increases the parameter count to around 80M. This lightweight design enables efficient training on a single P100 GPU without sacrificing spatial reasoning performance.

\subsection{Results and Ablation Study}

% \begin{table}[ht]
% \centering
% \begin{tabular}{cccc}
% \toprule
% \textbf{MoE} & \textbf{Phase 1} & \textbf{Phase 2} & \textbf{Score (\%)} \\
% \midrule
% \ding{55} & \ding{55} & \ding{51} & 63.65 \\
% \ding{55} & \ding{51} & \ding{51} & 65.09 \\
% \ding{51} & \ding{55} & \ding{51} & -- \\
% \ding{51} & \ding{51} & \ding{51} & -- \\
% \bottomrule
% \end{tabular}
% \caption{Ablation study of MoE and training phases. Phase 1 corresponds to free-form answers; Phase 2 corresponds to normalized answers. Scores are reported on the validation set.}
% \label{tab:ablation_phases}
% \end{table}

\begin{table}[ht]
    \centering
    \begin{tabular}{cccc}
    \toprule
    \textbf{MoE} & \textbf{Phase 1} & \textbf{Phase 2} & \textbf{Score (\%)} \\
    \midrule
    \ding{55} & \ding{51} & \ding{55} & 25.59 \\
    \ding{55} & \ding{55} & \ding{51} & 63.65 \\
    \ding{55} & \ding{51} & \ding{51} & 65.09 \\
    \ding{51} & \ding{55} & \ding{51} & 68.13 \\
    \ding{51} & \ding{51} & \ding{51} & \textbf{72.52} \\
    \bottomrule
    \end{tabular}
    \caption{Ablation study of training phases. Phase 1 corresponds to free-form answers; Phase 2 corresponds to normalized answers. Scores are reported on the validation set.}
    \label{tab:ablation_phases}
\end{table}
We performed an ablation study to assess the impact of the Mixture-of-Experts (MoE) module and the two-phase training strategy.
As shown in Table~\ref{tab:ablation_phases}, using only Phase 1 (free-form answer generation) yields the lowest performance (25.59\%).
Training solely on Phase 2 (normalized answers) substantially improves accuracy to 63.65\%, as it avoids post-processing errors from converting free-form outputs with an external instruction-tuned model (Qwen3 1.7B~\cite{qwen3}).
Enabling MoE in Phase 2 further boosts performance to 68.13\%, indicating that specialized experts are beneficial even without Phase 1 pretraining.

When both phases are combined, the model without MoE reaches 65.09\%, showing that free-form pretraining provides useful spatial priors for subsequent supervised normalization.
The highest score (72.52\%) is obtained when MoE is active across both phases, suggesting that expert-based fusion and structured normalization together yield complementary gains in spatial reasoning performance.

After the competition submission deadline, we extended the training duration for the non-MoE, two-phase configuration from 10 epochs to 25 epochs. When evaluated on the validation set, this setting achieved 88.52\% accuracy, compared to 65.09\% under the default epoch budget. Although this was not part of the official leaderboard submission, the result suggests that the model can further improve with longer training under relaxed compute constraints.

\subsection{Performance in the Challenge}
\begin{table}[ht]
\centering
\begin{tabular}{clr}
\toprule
\textbf{Rank} & \textbf{Team Name} & \textbf{Score (\%)} \\
\midrule
1 & UWIPL\_ETRI & 95.8638 \\
2 & HCMUT.VNU & 91.9735 \\
3 & Embia & 90.6772 \\
4 & MIZSU & 73.0606 \\
\textbf{5} & \textbf{HCMUS\_HTH} & \textbf{66.8861} \\
6 & MealsRetrieval & 53.4763 \\
7 & BKU22 & 50.3662 \\
8 & Smart Lab & 31.9245 \\
9 & AICV & 28.2993 \\
\bottomrule
\end{tabular}
\caption{Public leaderboard results on the Warehouse Spatial Intelligence track (Track 3, 9\textsuperscript{th} AI City Challenge 2025).}
\label{tab:leaderboard}
\end{table}

As shown in Table~\ref{tab:leaderboard}, TinyGiantVLM ranks 5$^{\text{th}}$ on the public leaderboard with a score of 66.89\%. This submission used an earlier model version without the MoE module. Despite a notable gap compared to top-performing entries (e.g., UWIPL\_ETRI at 95.86\%), our model was developed entirely under tight computational constraints, utilizing a single NVIDIA Tesla P100 GPU, a compact pretrained backbone, and lightweight fusion mechanisms. Our ablation study (Table~\ref{tab:ablation_phases}) isolates the contribution of the two-phase training strategy and highlights the model’s effectiveness under constrained resources. In addition to the overall leaderboard score, we further analyze performance across key spatial reasoning tasks. TinyGiantVLM achieves high accuracy on object counting (83.87\%) and left–right relation reasoning (98.40\%), demonstrating strong capabilities in symbolic and relational reasoning. Moderate performance on distance estimation (50.26\%) can be attributed to the inherent limitations of monocular depth and lack of explicit geometric supervision. The lowest score is observed on multiple-choice grounding (35.01\%), likely due to challenges in region-level visual disambiguation when confronted with multiple semantically similar candidates. These findings reveal the model’s strengths and weaknesses, guiding future improvements.
\section{Conclusion}
\label{sec:conclusion}

% In this paper, we present TinyGiantVLM, a lightweight visual language model designed for spatial reasoning in industrial warehouse environments. Our model integrates both global and region-level features from RGB and depth modalities, and adopts a modular architecture that supports cross-attention fusion and is designed to include a sparsely-gated Mixture-of-Experts (MoE) layer for task-specific reasoning. With only 64 million trainable parameters, TinyGiantVLM maintains a significantly smaller model footprint compared to top-performing systems, yet still ranked among the Top 5 teams on the PhysicalAI-Spatial-Intelligence-Warehouse dataset (Track 3, AI City Challenge 2025). This demonstrates the model’s ability to strike a favorable balance between accuracy and computational efficiency, an essential property for real-world industrial deployment under resource constraints. Future improvements may involve scaling to larger language backbones such as T5-Large~\cite{JMLR:v21:20-074} or LLaMA2~\cite{touvron2023llama2openfoundation}, learning dynamic region proposals, and exploring spatially grounded visual pretraining. Thanks to its lightweight and modular design, TinyGiantVLM provides a promising foundation for building adaptable and interpretable spatial reasoning systems with minimal supervision.

% \paragraph{Limitations.}
% The Mixture-of-Experts (MoE) module was not enabled due to runtime issues and will be evaluated in future updates, including the camera-ready version.

In this paper, we present TinyGiantVLM, a lightweight visual-language model for spatial reasoning in industrial warehouse environments. It integrates global and region-level features from RGB and depth modalities within a modular architecture supporting cross-attention fusion and optionally a sparsely-gated Mixture-of-Experts (MoE) layer for task-specific reasoning. The non-MoE variant, with only 64M trainable parameters, ranked in the Top 5 on the PhysicalAI-Spatial-Intelligence-Warehouse dataset (Track 3, AI City Challenge 2025 \cite{Tang25AICity25}), demonstrating a strong balance between accuracy and computational efficiency. Enabling MoE increases the model to ~80M parameters and yields further gains on the validation set in post-deadline ablation studies. With its compact, extensible design, TinyGiantVLM provides a promising foundation for adaptable and interpretable spatial reasoning under resource constraints.

\pagebreak
{
    \small
    \bibliographystyle{ieeenat_fullname}
    \bibliography{ref}
}


\end{document}